\documentclass[11pt,a4paper]{article}
\usepackage{hyperref}
\usepackage{acl2018}
\usepackage{times}
\usepackage{latexsym}
\usepackage{tikz}
\usetikzlibrary{positioning,shapes,shadows,arrows}

\usepackage{graphicx}
\PassOptionsToPackage{hyphens}{url}\usepackage{url}
\usepackage{listings}
\usepackage{color}
\usepackage{array,multirow}

\definecolor{codegreen}{rgb}{0,0.6,0}
\definecolor{codegray}{rgb}{0.5,0.5,0.5}
\definecolor{codepurple}{rgb}{0.58,0,0.82}
\definecolor{paperred}{rgb}{.84,0,0,}
\definecolor{greybackground}{rgb}{.98,0.98,0.98}

\lstdefinestyle{mystyle}{
    backgroundcolor=\color{greybackground},   
    commentstyle=\color{codegreen},
    keywordstyle=\color{purple},
    numberstyle=\tiny\color{codegray},
    stringstyle=\color{paperred},
    breakatwhitespace=false,         
    breaklines=true,                 
    captionpos=b,                    
    keepspaces=true,                 
    numbersep=5pt,                  
    showspaces=false,                
    showstringspaces=false,
    showtabs=false,                  
    tabsize=2,frame=single,
    basicstyle=\scriptsize\ttfamily,
    belowskip=-1.5em,
}
\usepackage{subfig}
\aclfinalcopy 

\title{Universal Sentence Encoder}

\author{Daniel Cer\textsuperscript{$a$}, 
        Yinfei Yang\textsuperscript{$a$}, 
        Sheng-yi Kong\textsuperscript{$a$},
        \rm\textbf{Nan Hua\textsuperscript{$a$}, 
        Nicole Limtiaco\textsuperscript{$b$},} \\
        \rm\textbf{Rhomni St. John\textsuperscript{$a$},
        Noah Constant\textsuperscript{$a$},
        Mario Guajardo-C\'espedes\textsuperscript{$a$},
        Steve Yuan\textsuperscript{$c$},} \\
\rm\textbf{Chris Tar\textsuperscript{$a$}, Yun{-}Hsuan Sung\textsuperscript{$a$},
        Brian Strope\textsuperscript{$a$},
        Ray Kurzweil\textsuperscript{$a$}} \AND
  {\rm\textsuperscript{$a$}Google Research}\\Mountain View, CA \And
  {\rm\textsuperscript{$b$}Google Research}\\New York, NY
  \And
  {\rm\textsuperscript{$c$}Google}\\Cambridge, MA
}

\date{}

\begin{document}
\maketitle
\begin{abstract}
We present models for encoding sentences into embedding vectors that specifically target transfer learning to other NLP tasks. The models are efficient and result in accurate performance on diverse transfer tasks. Two variants of the encoding models allow for trade-offs between accuracy and compute resources. For both variants, we investigate and report the relationship between model complexity, resource consumption, the availability of transfer task training data, and task performance. Comparisons are made with baselines that use word level transfer learning via pretrained word embeddings as well as baselines do not use any transfer learning. We find that transfer learning using sentence embeddings tends to outperform word level transfer. With transfer learning via sentence embeddings, we observe surprisingly good performance with minimal amounts of supervised training data for a transfer task. We obtain encouraging results on Word Embedding Association Tests (WEAT) targeted at detecting model bias. Our pre-trained sentence encoding models are made freely available for download and on TF Hub.

\end{abstract}

\section{Introduction}

Limited amounts of training data are available for many NLP tasks. This presents a challenge for data hungry deep learning methods. Given the high cost of annotating supervised training data, very large training sets are usually not available for most research or industry NLP tasks. Many models address the problem by implicitly performing limited transfer learning through the use of pre-trained word embeddings such as those produced by word2vec \cite{mikolov2013} or GloVe \cite{pennington2014glove}. However, recent work has demonstrated strong transfer task performance using pre-trained sentence level embeddings \cite{conneau2017}.

\begin{figure}
  \includegraphics[width=7.5cm]{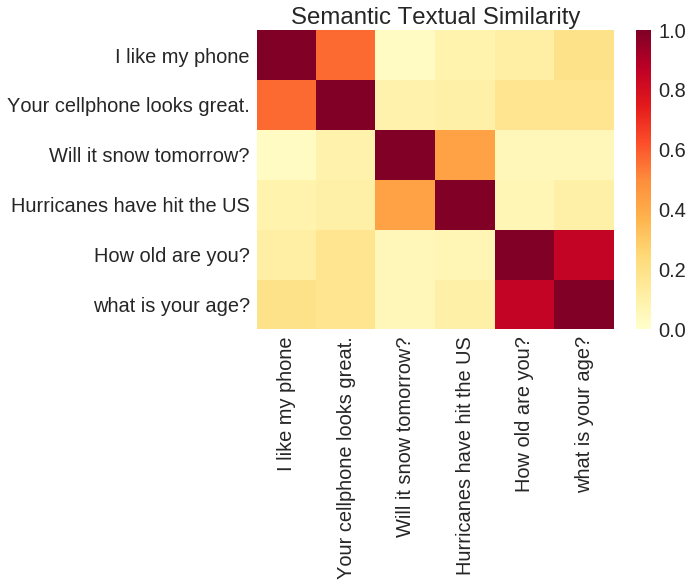}
  \caption{Sentence similarity scores using embeddings from the universal sentence encoder.}
\label{fig:sts_vis}
\end{figure}

In this paper, we present two models for producing sentence embeddings that demonstrate good transfer to a number of other of other NLP tasks. We include experiments with varying amounts of transfer task training data to illustrate the relationship between transfer task performance and training set size. We find that our sentence embeddings can be used to obtain surprisingly good task performance with remarkably little task specific training data. The sentence encoding models are made publicly available on TF Hub. 

Engineering characteristics of models used for transfer learning are an important consideration. We discuss modeling trade-offs regarding memory requirements as well as compute time on CPU and GPU. Resource consumption comparisons are made for sentences of varying lengths.

\section{Model Toolkit}

We make available two new models for encoding sentences into embedding vectors. One makes use of the transformer \cite{vaswani2017} architecture, while the other is formulated as a deep averaging network (DAN) \cite{iyyer2015}. Both models are implemented in TensorFlow \cite{abadi2016} and are available to download from TF Hub:\footnote{The encoding model for the DAN based encoder is already available. The transformer based encoder will be made available at a later point.} 

\begin{center}
\footnotesize
\texttt{\href{https://tfhub.dev/google/universal-sentence-encoder/1}
       {https://tfhub.dev/google/} \\
       \href{https://tfhub.dev/google/universal-sentence-encoder/1}
       {universal-sentence-encoder/1}}
\end{center}

The models take as input English strings and produce as output a fixed dimensional embedding representation of the string. Listing \ref{uesnippet} provides a minimal code snippet to convert a sentence into a tensor containing its sentence embedding. The {\tt embedding} tensor can be used directly or incorporated into larger model graphs for specific tasks.\footnote{Visit \href{https://colab.research.google.com/}{https://colab.research.google.com/} to try the code snippet in Listing \ref{uesnippet}. Example code and documentation is available on the universal encoder website provided above.}  

As illustrated in Figure \ref{fig:sts_vis}, the sentence embeddings can be trivially used to compute sentence level semantic similarity scores that achieve excellent performance on the semantic textual similarity (STS) Benchmark \cite{cer2017}. When included within larger models, the sentence encoding models can be fine tuned for specific tasks using gradient based updates.

\section{Encoders}
We introduce the model architecture for our two encoding models in this section. Our two encoders have different design goals. One based on the transformer architecture targets high accuracy at the cost of greater model complexity and resource consumption. The other targets efficient inference with slightly reduced accuracy.

\lstset{framesep=5pt}
\begin{lstlisting}[language=Python,label=uesnippet,float,caption=Python example code for using the universal sentence encoder.]
import tensorflow_hub as hub

embed = hub.Module("https://tfhub.dev/google/"
    "universal-sentence-encoder/1")
    
embedding = embed([
    "The quick brown fox jumps over the lazy dog."])
\end{lstlisting}

\subsection{Transformer}

The transformer based sentence encoding model constructs sentence embeddings using the encoding sub-graph of the transformer architecture \cite{vaswani2017}. This sub-graph uses attention to compute context aware representations of words in a sentence that take into account both the ordering and identity of all the other words. The context aware word representations are converted to a fixed length sentence encoding vector by computing the element-wise sum of the representations at each word position.\footnote{We then divide by the square root of the length of the sentence so that the differences between short sentences are not dominated by sentence length effects} The encoder takes as input a lowercased PTB tokenized string and outputs a 512 dimensional vector as the sentence embedding. 

The encoding model is designed to be as general purpose as possible. This is accomplished by using multi-task learning whereby a single encoding model is used to feed multiple downstream tasks. The supported tasks include: a Skip-Thought like task \cite{kiros2015} for the unsupervised learning from arbitrary running text; a conversational input-response task for the inclusion of parsed conversational data \cite{henderson2017}; and classification tasks for training on supervised data. The Skip-Thought task replaces the LSTM \cite{hochreiter1997} used in the original formulation with a model based on the Transformer architecture.

As will be shown in the experimental results below, the transformer based encoder achieves the best overall transfer task performance. However, this comes at the cost of compute time and memory usage scaling dramatically with sentence length.

\subsection{Deep Averaging Network (DAN)}

The second encoding model makes use of a deep averaging network (DAN) \cite{iyyer2015} whereby input embeddings for words and bi-grams are first averaged together and then passed through a feedforward deep neural network (DNN) to produce sentence embeddings.  
Similar to the Transformer encoder, the DAN encoder takes as input a lowercased PTB tokenized string and outputs a 512 dimensional sentence embedding. The DAN encoder is trained similarly to the Transformer based encoder. We make use of multitask learning whereby a single DAN encoder is used to supply sentence embeddings for multiple downstream tasks.

The primary advantage of the DAN encoder is that compute time is linear in the length of the input sequence. Similar to \newcite{iyyer2015}, our results demonstrate that DANs achieve strong baseline performance on text classification tasks.

\subsection{Encoder Training Data}

Unsupervised training data for the sentence encoding models are drawn from a variety of web sources. The sources are Wikipedia, web news, web question-answer pages and discussion forums. We augment unsupervised learning with training on supervised data from the Stanford Natural Language Inference (SNLI) corpus \cite{bowman2015}. Similar to the findings of \newcite{conneau2017}, we observe that training to SNLI improves transfer performance.

\section{Transfer Tasks}

This section presents an overview of the data used for the transfer learning experiments and the Word Embedding Association Test (WEAT) data used to characterize model bias.\footnote{For the datasets MR, CR, and SUBJ, SST, and TREC we use the preparation of the data provided by \newcite{conneau2017}.} Table \ref{tab:eval-set-size} summarizes the number of samples provided by the test portion of each evaluation set and, when available, the size of the dev and training data.


\paragraph{MR}: Movie review snippet sentiment on a five star scale \cite{pang2005}.

\paragraph{CR}: Sentiment of sentences mined from customer reviews \cite{hu2004}.  

\paragraph{SUBJ}: Subjectivity of sentences from movie reviews and plot summaries \cite{pang2004}.

\paragraph{MPQA}: Phrase level opinion polarity from news data \cite{wiebe2005}.

\paragraph{TREC}: Fine grained question classification sourced from TREC \cite{li2002}.

\paragraph{SST}:  Binary phrase level sentiment classification \cite{socher2013}.

\paragraph{STS Benchmark}: Semantic textual similarity (STS) between sentence pairs scored by Pearson correlation with human judgments \cite{cer2017}.

\paragraph{WEAT}: Word pairs from the psychology literature on implicit association tests (IAT) that are used to characterize model bias \cite{caliskan183}.

\begin{table}[ht]
\small
\begin{center}

\begin{tabular}{|c|c|c|c|}
\hline
Dataset & Train & Dev & Test \\
\hline
SST       & 67,349 & 872 & 1,821 \\
STS Bench &  5,749 & 1,500 & 1,379 \\
\hline
TREC      & 5,452 &  -   &   500 \\
\hline
MR        & - & - & 10,662 \\
CR        & - & - & 3,775 \\
SUBJ      & - & - & 10,000 \\
MPQA      & - & - & 10,606 \\
\hline
\end{tabular}

\end{center}
\caption{Transfer task evaluation sets}
\label{tab:eval-set-size}
\end{table}

\begin{table*}[ht!]
\small
\begin{center}

\begin{tabular}{|c|c|c|c|c|c|c|c|}
\hline
\multirow{2}{*}{Model} & \multirow{2}{*}{MR} & \multirow{2}{*}{CR} & \multirow{2}{*}{SUBJ} & \multirow{2}{*}{MPQA} & \multirow{2}{*}{TREC} & \multirow{2}{*}{SST} & STS Bench \\
& & & & & & & (dev / test) \\
\hline
\multicolumn{8}{|c|}{Sentence \& Word Embedding Transfer Learning} \\
\hline
USE\_D+DAN (w2v w.e.)  & 77.11  & 81.71  & 93.12  & 87.01  & 94.72  & 82.14 & -- \\
USE\_D+CNN (w2v w.e.)  & 78.20  & 82.04  & 93.24  & 85.87  & 97.67  & 85.29 & -- \\
USE\_T+DAN (w2v w.e.)  & 81.32  & 86.66  & 93.90  & 88.14  & 95.51  & 86.62 & -- \\
USE\_T+CNN (w2v w.e.)  & 81.18  & 87.45  & 93.58  & 87.32  & 98.07  & 86.69 & -- \\
\hline
\multicolumn{8}{|c|}{\emph{Sentence Embedding Transfer Learning}} \\
\hline
USE\_D  & 74.45  & 80.97  & 92.65  & 85.38  & 91.19  & 77.62 & 0.763 / 0.719 (r) \\
USE\_T  & 81.44  & 87.43  & 93.87  & 86.98  & 92.51  & 85.38 & 0.814 / 0.782 (r) \\
USE\_D+DAN (lrn w.e.)  & 77.57  & 81.93  & 92.91  & 85.97  & 95.86  & 83.41 & --  \\
USE\_D+CNN (lrn w.e.)  & 78.49  & 81.49  & 92.99  & 85.53  & 97.71  & 85.27 & --  \\
USE\_T+DAN (lrn w.e.)  & 81.36  & 86.08  & 93.66  & 87.14  & 96.60  & 86.24 & --  \\
USE\_T+CNN (lrn w.e.)  & 81.59  & 86.45  & 93.36  & 86.85  & 97.44  & 87.21 & --  \\
\hline
\multicolumn{8}{|c|}{\emph{Word Embedding Transfer Learning}} \\
\hline
DAN (w2v w.e.)  & 74.75  & 75.24  & 90.80  & 81.25  & 85.69  &  80.24 & -- \\
CNN (w2v w.e.)  & 75.10  & 80.18  & 90.84  & 81.38  & 97.32  &  83.74 & -- \\
\hline
\multicolumn{8}{|c|}{\emph{Baselines with No Transfer Learning}} \\
\hline
DAN (lrn w.e.)  & 75.97  & 76.91  & 89.49  & 80.93  & 93.88  &  81.52 & -- \\
CNN (lrn w.e.)  & 76.39  & 79.39  & 91.18  & 82.20  & 95.82  &  84.90 & -- \\
\hline
\end{tabular}
\end{center}
\caption{Model performance on transfer tasks. \emph{USE\_T} is the universal sentence encoder (USE) using Transformer. \emph{USE\_D} is the universal encoder DAN model. Models tagged with \emph{w2v w.e.}\ make use of pre-training word2vec skip-gram embeddings for the transfer task model, while models tagged with \emph{lrn w.e.}\ use randomly initialized word embeddings that are learned only on the transfer task data. Accuracy is reported for all evaluations except STS Bench where we report the Pearson correlation of the similarity scores with human judgments. Pairwise similarity scores are computed directly using the sentence embeddings from the universal sentence encoder as in Eq. (1).
}
\label{tab:trans-model-performance}
\end{table*}

\section{Transfer Learning Models}

For sentence classification transfer tasks, the output of the transformer and DAN sentence encoders are provided to a task specific DNN. For the pairwise semantic similarity task, we directly assess the similarity of the sentence embeddings produced by our two encoders. As shown Eq. \ref{eqsimscore}, we first compute the cosine similarity of the two sentence embeddings and then use arccos to convert the cosine similarity into an angular distance.\footnote{We find that using a similarity based on angular distance performs better on average than raw cosine similarity.}

\begin{equation}
  \textrm{sim}({\bf u}, {\bf v}) = \Big(1 - \arccos\bigg(\frac{{\bf u} \cdot {\bf v}}{||{\bf u}|| ~ ||{\bf v}||}\bigg) / \pi\Big)
\label{eqsimscore}
\end{equation}

\subsection{Baselines}

For each transfer task, we include baselines that only make use of word level transfer and baselines that make use of no transfer learning at all. For word level transfer, we use word embeddings from a word2vec skip-gram model trained on a corpus of news data \cite{mikolov2013}. The pretrained word embeddings are included as input to two model types: a convolutional neural network models (CNN) \cite{kim2014}; a DAN. The baselines that use pretrained word embeddings allow us to contrast word versus sentence level transfer. Additional baseline CNN and DAN models are trained without using any pretrained word or sentence embeddings.

\subsection{Combined Transfer Models}

We explore combining the sentence and word level transfer models by concatenating their representations prior to feeding the combined representation to the transfer task classification layers. For completeness, we also explore concatenating the representations from sentence level transfer models with the baseline models that do not make use of word level transfer learning.

\section{Experiments}
Transfer task model hyperparamaters are tuned using a combination of Vizier \cite{golovin2017} and light manual tuning. When available, model hyperparameters are tuned using task dev sets. Otherwise, hyperparameters are tuned by cross-validation on the task training data when available or the evaluation test data when neither training nor dev data are provided. Training repeats ten times for each transfer task model with different randomly initialized weights and we report evaluation results by averaging across runs.

Transfer learning is critically important when training data for a target task is limited. We explore the impact on task performance of varying the amount of training data available for the task both with and without the use of transfer learning. Contrasting the transformer and DAN based encoders, we demonstrate trade-offs in model complexity and the amount of data required to reach a desired level of accuracy on a task.

To assess bias in our encoding models, we evaluate the strength of various associations learned by our model on WEAT word lists. We compare our result to those of \newcite{caliskan183} who discovered that word embeddings could be used to reproduce human performance on implicit association tasks for both benign and potentially undesirable associations. 

\begin{table*}
\begin{tabular}{|c|c|c|c|c|c|c|c|}
\hline
Model & SST 1k & SST 2k & SST 4k & SST 8k & SST 16k & SST 32k & SST 67.3k \\
\hline
\multicolumn{8}{|c|}{Sentence \& Word Embedding Transfer Learning} \\
\hline
USE\_D+DNN (w2v w.e.)  & 78.65  & 78.68  & 79.07  & 81.69  & 81.14  & 81.47 & 82.14 \\
USE\_D+CNN (w2v w.e.)  & 77.79  & 79.19  & 79.75  & 82.32  & 82.70  & 83.56 & 85.29 \\
USE\_T+DNN (w2v w.e.)  & 85.24  & 84.75  & 85.05  & 86.48  & 86.44  & 86.38 & 86.62 \\
USE\_T+CNN (w2v w.e.)  & 84.44  & 84.16  & 84.77  & 85.70  & 85.22  & 86.38 & 86.69 \\
\hline
\multicolumn{8}{|c|}{\emph{Sentence Embedding Transfer Learning}} \\
\hline
USE\_D  & 77.47  & 76.38  & 77.39  & 79.02  & 78.38  & 77.79 & 77.62 \\
USE\_T  & 84.85  & 84.25  & 85.18  & 85.63  & 85.83  & 85.59 & 85.38 \\
USE\_D+DNN (lrn w.e.)  & 75.90  & 78.68  & 79.01  & 82.31  & 82.31  & 82.14 & 83.41 \\
USE\_D+CNN (lrn w.e.)  & 77.28  & 77.74  & 79.84  & 81.83  & 82.64  & 84.24 & 85.27 \\
USE\_T+DNN (lrn w.e.)  & 84.51  & 84.87  & 84.55  & 85.96  & 85.62  & 85.86 & 86.24 \\
USE\_T+CNN (lrn w.e.)  & 82.66  & 83.73  & 84.23  & 85.74  & 86.06  & 86.97 & 87.21 \\
\hline
\multicolumn{8}{|c|}{\emph{Word Embedding Transfer Learning}} \\
\hline
DNN (w2v w.e.)  & 66.34  & 69.67  & 73.03  & 77.42  & 78.29  & 79.81 & 80.24 \\
CNN (w2v w.e.)  & 68.10  & 71.80  & 74.91  & 78.86  & 80.83  & 81.98 & 83.74 \\
\hline
\multicolumn{8}{|c|}{\emph{Baselines with No Transfer Learning}} \\
\hline
DNN (lrn w.e.)  & 66.87  & 71.23  & 73.70  & 77.85  & 78.07  & 80.15 & 81.52 \\
CNN (lrn w.e.)  & 67.98  & 71.81  & 74.90  & 79.14  & 81.04  & 82.72 & 84.90 \\
\hline
\end{tabular}
\caption{Task performance on SST for varying amounts of training data. SST 67.3k represents the full training set. Using only 1,000 examples for training, transfer learning from USE\_T is able to obtain performance that rivals many of the other models trained on the full 67.3 thousand example training set. }
\label{table:transfer_perf_vs_train_set_sz}
\end{table*}

\section{Results}

Transfer task performance is summarized in Table \ref{tab:trans-model-performance}. We observe that transfer learning from the transformer based sentence encoder usually performs as good or better than transfer learning from the DAN encoder. Hoewver, transfer learning using the simpler and fast DAN encoder can for some tasks perform as well or better than the more sophisticated transformer encoder. Models that make use of sentence level transfer learning tend to perform better than models that only use word level transfer. The best performance on most tasks is obtained by models that make use of both sentence and word level transfer.

Table \ref{table:transfer_perf_vs_train_set_sz} illustrates transfer task performance for varying amounts of training data. We observe that, for smaller quantities of data, sentence level transfer learning can achieve surprisingly good task performance. As the training set size increases, models that do not make use of transfer learning approach the performance of the other models.

\begin{figure*}[ht!]
\small
     \subfloat[CPU Time vs.\ Sentence Length\label{fig:cputime}]{
       \includegraphics[width=4.5cm]{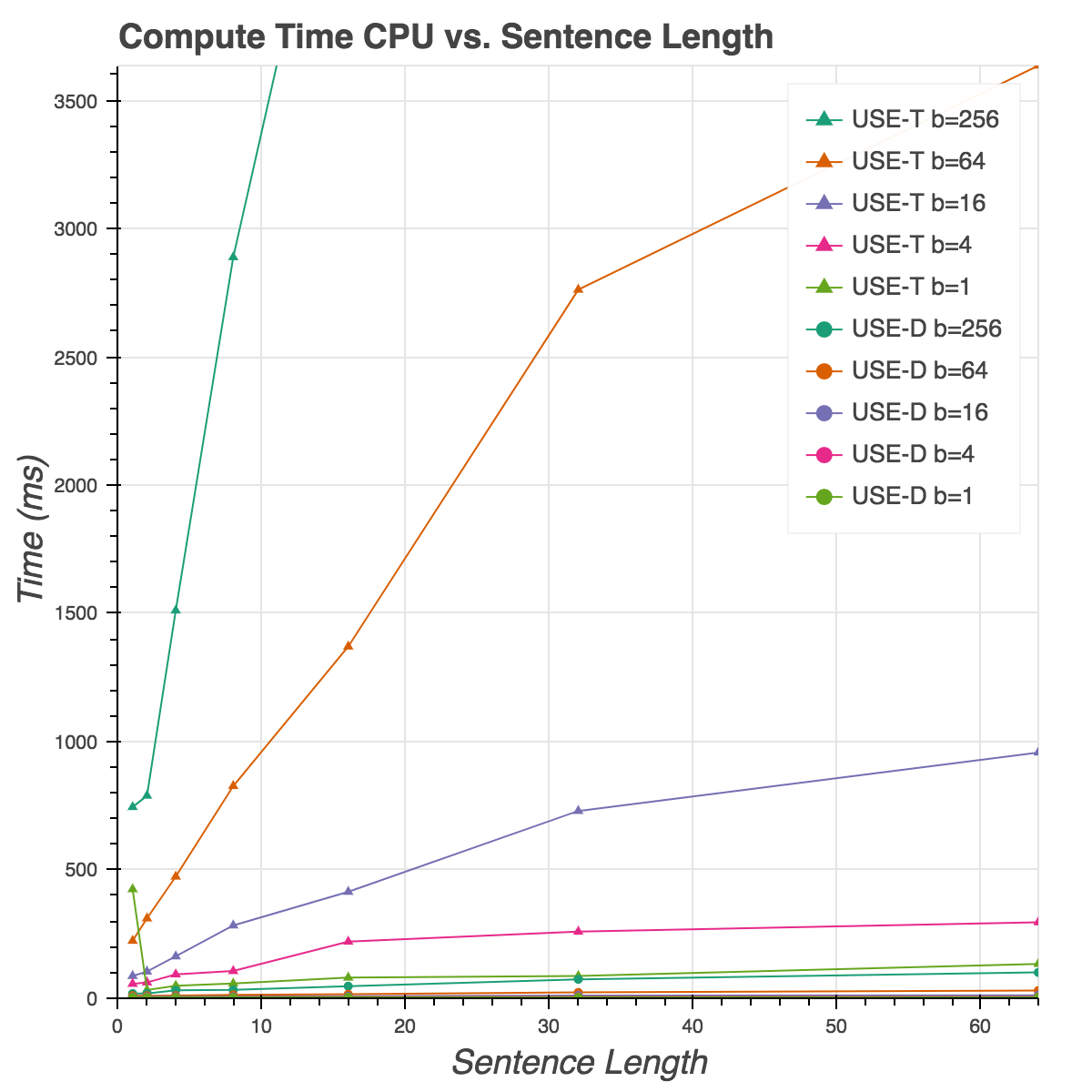}
     }
     \hfill
     \subfloat[GPU Time vs.\ Sentence Length\label{fig:gputime}]{
       \includegraphics[width=4.5cm]{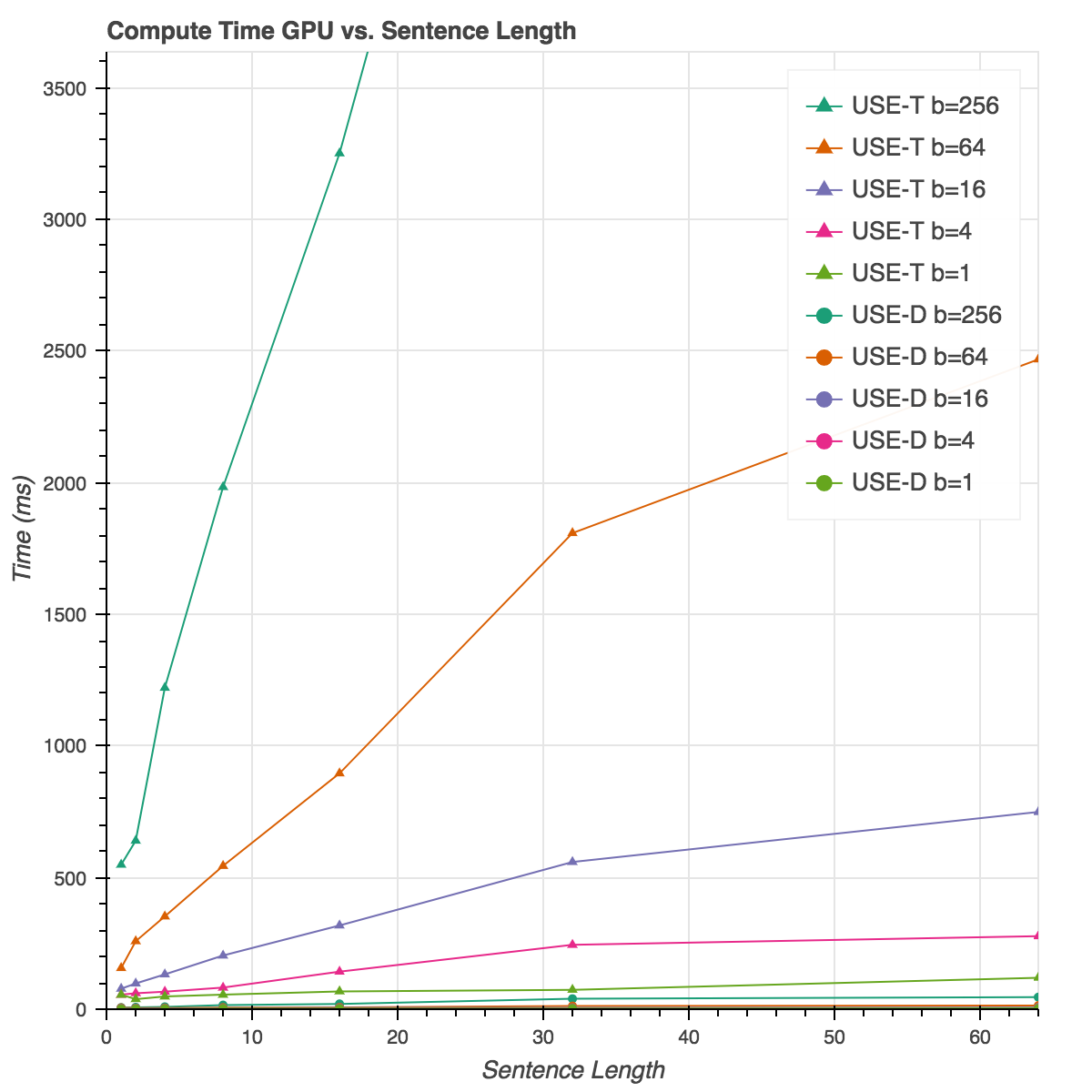}

     }
     \hfill
     \subfloat[Memory vs.\ Sentence Length\label{fig:modelvshuman:en-es}]{
       \includegraphics[width=4.5cm]{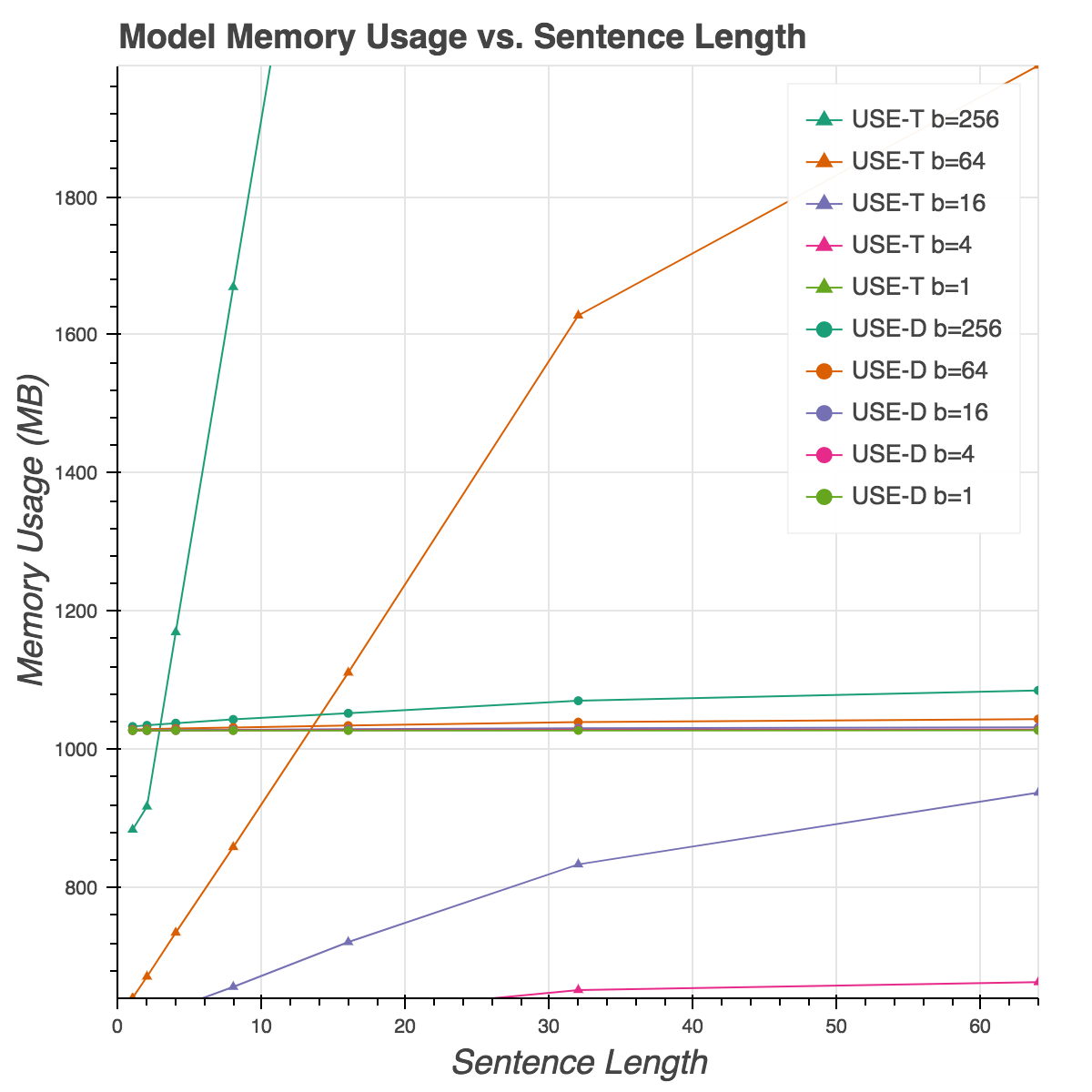}
     }
\caption{Model Resource Usage for both USE\_D and USE\_T at different batch sizes and sentence lengths.  }
\label{fig:engbench}
\end{figure*}

Table \ref{tab:caliskancomp} contrasts \newcite{caliskan183}'s findings on bias within GloVe embeddings with the DAN variant of the universal encoder. Similar to GloVe, our model reproduces human associations between flowers vs. insects and pleasantness vs. unpleasantness. However, our model demonstrates weaker associations than GloVe for probes targeted at revealing at ageism, racism and sexism.\footnote{Researchers and developers are strongly encouraged to independently verify whether biases in their overall model or model components impacts their use case. For resources on ML fairness visit \href{https://developers.google.com/machine-learning/fairness-overview/}{https://developers.google.com/machine-learning/fairness-overview/}. } The differences in word association patterns can be attributed to differences in the training data composition and the mixture of tasks used to train the sentence embeddings.

\begin{table*}[ht!]
\footnotesize
\begin{center}

\begin{tabular}{|c|c|c|c|c|c|c|}
\hline
\multirow{2}{*}{Target words} & \multirow{2}{*}{Attrib. words} & \multirow{2}{*}{Ref} &
  \multicolumn{2}{|c|}{GloVe} &
  \multicolumn{2}{|c|}{ Uni. Enc. (DAN) } \\\cline{4-7}
  & & & d & p & d & p \\
\hline
Eur.-American
vs Afr.-American
names & Pleasant vs.\ 
Unpleasant 1 & $a$ & 1.41 & $10^{-8}$ & 0.361 & 0.035 \\
\hline
Eur.-American
vs.\ Afr.-American
names & Pleasant vs.
Unpleasant from (a) & $b$ & 1.50 & $10^{-4}$ & -0.372 & 0.87 \\
\hline
Eur.-American
vs.\ Afr.-American
names & Pleasant vs.
Unpleasant from $(c)$ & $b$ & 1.28 & $10^{-3}$ & 0.721 & 0.015 \\
\hline
Male vs.\ female
names & Career vs
family & $c$ & 1.81 & $10^{-3}$ & 0.0248 & 0.48 \\
\hline
Math vs.\ arts &
Male vs.\
female terms & $c$ & 1.06 & 0.018 & 0.588 & 0.12 \\
\hline
Science vs.\ arts & Male vs
female terms & $d$ & 1.24 & $10^{-2}$ & 0.236 & 0.32 \\
\hline
Mental vs.\
physical disease & Temporary vs
permanent & $e$ & 1.38 & $10^{-2}$ & 1.60 & 0.0027 \\
\hline
Young vs old
people’s names & Pleasant vs
unpleasant & $c$ & 1.21 & $10^{-2}$ & 1.01 & 0.022 \\
\hline
\hline
Flowers vs.\
insects & 
Pleasant vs.\ Unpleasant & $a$ & 1.50 & $10^{-7}$ & 1.38 & $10^{-7}$ \\
\hline
Instruments vs.\ Weapons &
Pleasant vs
Unpleasant & $a$ & 1.53 & $10^{-7}$ & 1.44 & $10^{-7}$  \\ 
\hline
\end{tabular}
\end{center}
\caption{Word Embedding Association Tests (WEAT) for GloVe and the Universal Encoder. Effect size is reported as Cohen's d over the mean cosine similarity scores across grouped attribute words. Statistical significance is reported for 1 tailed p-scores. The letters in the \emph{Ref} column indicates the source of the IAT word lists: $(a)$ \newcite{greenwald1998} $(b)$ \newcite{bertrand2004} $(c)$ \newcite{nosek2002a} $(d)$ \newcite{nosek2002b} $(e)$ \newcite{monteith2011}.}
\label{tab:caliskancomp}
\end{table*}

\subsection{Discussion}

Transfer learning leads to performance improvements on many tasks. Using transfer learning is more critical when less training data is available. When task performance is close, the correct modeling choice should take into account engineering trade-offs regarding the memory and compute resource requirements introduced by the different models that could be used.  

\section{Resource Usage}

This section describes memory and compute resource usage for the transformer and DAN sentence encoding models for different sentence lengths. Figure \ref{fig:engbench} plots model resource usage against sentence length. 

\paragraph{Compute Usage} The transformer model time complexity is $O(n^2)$ in sentence length, while the DAN model is $O(n)$. As seen in Figure \ref{fig:engbench} (a-b), for short sentences, the transformer encoding model is only moderately slower than the much simpler DAN model. However, compute time for transformer increases noticeably as sentence length increases. In contrast, the compute time for the DAN model stays nearly constant as sentence length is increased. Since the DAN model is remarkably computational efficient, using GPUs over CPUs will often have a much larger practical impact for the transformer based encoder. 

\paragraph{Memory Usage}

The transformer model space complexity also scales quadratically, $O(n^2)$, in sentence length, while the DAN model space complexity is constant in the length of the sentence. Similar to compute usage, memory usage for the transformer model increases quickly with sentence length, while the memory usage for the DAN model remains constant. We note that, for the DAN model, memory usage is dominated by the parameters used to store the model unigram and bigram embeddings. Since the transformer model only needs to store unigram embeddings, for short sequences it requires nearly half as much memory as the DAN model.

\section{Conclusion}

Both the transformer and DAN based universal encoding models provide sentence level embeddings that demonstrate strong transfer performance on a number of NLP tasks. The sentence level embeddings surpass the performance of transfer learning using word level embeddings alone. Models that make use of sentence and word level transfer achieve the best overall performance. We observe that transfer learning is most helpful when limited training data is available for the transfer task. The encoding models make different trade-offs regarding accuracy and model complexity that should be considered when choosing the best model for a particular application. The pre-trained encoding models will be made publicly available for research and use in applications that can benefit from a better understanding of natural language.

\section*{Acknowledgments}

We thank our teammates from Descartes, Ai.h and other Google groups for their feedback and suggestions. Special thanks goes to Ben Packer and Yoni Halpern for implementing the WEAT assessments and discussions on model bias.

\bibliography{acl2018}
\bibliographystyle{acl_natbib}

\appendix

\end{document}